\pdfoutput=1

\documentclass[10pt, a4paper]{article}

\usepackage{lrec-coling2024} % this is the new style
\usepackage{booktabs, multirow} % for borders and merged ranges
\usepackage{soul}
\usepackage{graphicx}
\usepackage{tabularx}
\usepackage{afterpage}

\title{CamemBERT-bio: Leveraging Continual Pre-training for Cost-Effective Models on French Biomedical Data\\ \vspace*{.5\baselineskip}}

\name{Rian Touchent, Laurent Romary, Eric de La Clergerie} 

\address{Inria, Sorbonne Université \\
         2 rue Simone IFF 75012 Paris, 21 rue de l'école de médecine
75006 Paris \\
         \{rian.touchent,laurent.romary,eric.de\_la\_clergerie\}@inria.fr\\}

\abstract{
Clinical data in hospitals are increasingly accessible for research through clinical data warehouses. However these documents are unstructured and it is therefore necessary to extract information from medical reports to conduct clinical studies. Transfer learning with BERT-like models such as CamemBERT has allowed major advances for French, especially for named entity recognition. However, these models are trained for plain language and are less efficient on biomedical data. Addressing this gap, we introduce CamemBERT-bio, a dedicated French biomedical model derived from a new public French biomedical dataset. Through continual pre-training of the original CamemBERT, CamemBERT-bio achieves an improvement of 2.54 points of F1-score on average across various biomedical named entity recognition tasks, reinforcing the potential of continual pre-training as an equally proficient yet less computationally intensive alternative to training from scratch. Additionally, we highlight the importance of using a standard evaluation protocol that provides a clear view of the current state-of-the-art for French biomedical models.
 \\ \newline  }

\begin{document}

\maketitleabstract

\section{Introduction}

In recent years, there has been a development of clinical data warehouses (CDWs) in hospitals. These are clinical databases aimed at being more accessible for research purposes. These documents represent an opportunity for massive clinical studies using real data. They can take various forms within Electronic Health Records (EHR), such as reports, medical imaging, or prescriptions. However, most of the information is found in clinical reports. It is estimated that up to 80\% of entities are missing from other modalities \cite{raghavan_how_2014}. Although these data are highly valuable, they are unstructured, which requires preprocessing before they can be used in a clinical study.

BERT-based models \citeplanguageresource{devlin_bert_2019} consistently demonstrate state-of-the-art results for a wide range of natural language processing tasks. The adaptation of BERT to the French language, particularly with the CamemBERT model \citeplanguageresource{martin-etal-2020-camembert}, has replicated these performances in French natural language processing. CamemBERT is based on RoBERTa \citeplanguageresource{liu_roberta_2019}, which is a more efficient version of BERT. It is trained on a French corpus extracted from the web called OSCAR \citeplanguageresource{OrtizSuarezSagotRomary2019}.

To extract information from medical reports, it is necessary to have high-performing language models trained on French clinical data, particularly for named entity recognition. It is possible to simply use CamemBERT; however, the results of this model on biomedical data are disappointing \cite{cardon_presentation_2020}, as it exhibits lower performance compared to heuristic models on certain evaluation datasets. These results are predictable because CamemBERT is trained on plain language, often sourced from web pages such as forums. However, biomedical data, especially clinical data, are significantly different. They contain technical terms that are very rare or absent in everyday language, and they have a radically distinct style, often telegraphic, rarely consisting of complete sentences, with varying abbreviations.

One of the major challenges with healthcare data warehouses is data confidentiality. These data are regulated and subject to strong regulations by the CNIL (French Data Protection Authority). As a result, adaptations of CamemBERT to the biomedical domain conducted within hospital infrastructures \cite{dura_learning_2022} cannot be publicly released. Their training datasets are subject to publication constraints. These constraints also apply to the resulting models. Therefore, it is not possible to exchange these models between different healthcare institutions. A publicly available model would not have these constraints and could be used in various institutions.

Using continual-pretraining on a new French biomedical corpus, we introduce a new model named CamemBERT-bio, which shows a 2.54 points improvement in F-score on several French biomedical named entity recognition tasks. Furthermore, we engage in a discussion on the evaluation of French clinical models, emphasizing the importance of adhering to established standard practices. By following this methodology, we successfully showcase the effectiveness of continual-pretraining in the context of a French model, which contrast with recent suggestions pertaining to the same domain and language \citeplanguageresource{labrak2023drbert}. \\

In this article, we present three main contributions :%\footnote{Our contributions are available on the Hugging Face hub: \href{https://hf.co/almanach/camembert-bio-base}{almanach/camembert-bio-base}}
\begin{itemize}
\item The creation of a new public French dataset specialized in the biomedical domain.
\item The introduction of a publicly available adaptation of CamemBERT for the biomedical domain, which demonstrates improved performance on named entity recognition tasks.
\item The demonstration that continual-pretraining on a French model is successful, necessitating a reevaluation of previous work due to the impact of evaluation methodology on result interpretation.
\end{itemize}

\section{Related Works}

Research on adapting language models to new domains is extensive. \citet{gururangan_dont_2020} demonstrate that a second phase of pre-training on a target domain can improve performance on various tasks, even when the target domain corpus is small in size. In the biomedical domain, it has been observed that there can be up to a 3-point increase in F-measure compared to the same model without the second phase of pre-training.

This study by \citet{gururangan_dont_2020} has inspired the creation of new models based on BERT, utilizing a second phase of pre-training on various specialized domains. \citetlanguageresource{lee_biobert_2019} introduced BioBERT, a BERT-based model specialized for biomedical text in English. BioBERT demonstrates improved performance on various biomedical NLP tasks, including a $0.62\%$ F-measure improvement on named entity recognition, a $2.80\%$ F-measure improvement on relation extraction, and a $12.24\%$ MMR improvement on question-answering. The second phase of pre-training is conducted on a corpus extracted from PubMed and PMC, consisting of approximately 18 billion words from biomedical scientific articles. While the corpus is substantial and solely composed of scientific-style text, performance gains are observed across all text styles. The presence of medical vocabulary in the corpus likely contributes to significant improvement compared to general language models.

Training new models from scratch is also a viable approach. This is explored in SciBERT \citeplanguageresource{beltagy_scibert_2019} and PubMedBERT \citeplanguageresource{gu_domain-specific_2022}, two models specialized in biomedical scientific articles. PubMedBERT demonstrates that this method yields better performance than models trained with a second phase of specialization. However, the performance gains are relatively modest, and this approach is more computationally expensive. Starting from scratch requires longer training times and a larger corpus to achieve comparable performance.

For the French language, the reference models are CamemBERT \citeplanguageresource{martin-etal-2020-camembert} and FlauBERT \citeplanguageresource{le2020flaubert}. Several works have attempted to adapt CamemBERT to the biomedical domain. 

\citet{copara_contextualized_2020} explored a second phase of pre-training on 31,000 French biomedical scientific articles. However, they did not observe a significant improvement on a clinical named entity recognition task using the large version of CamemBERT. This could be explained by the combination of a relatively small corpus (31k documents compared to the 18 billion words in BioBERT) and the large version of the CamemBERT model.

\citet{le_clercq_de_lannoy_strategies_2022} also adapted CamemBERT to the biomedical domain. They aggregated documents from various sources, including PubMed, Cochrane, ISTEX, and Wikipedia, forming a larger partially public corpus of approximately 136 million words. They observed a 2-point improvement in F-measure on a named entity recognition evaluation set composed of drug notices (EMEA), but no significant improvement on a set composed of scientific article titles (MEDLINE).

\citet{dura_learning_2022} continued the pre-training of CamemBERT on 21 million clinical documents from the APHP (Assistance Publique - Hôpitaux de Paris) clinical data warehouse. They observed a significant 3\% improvement on APMed, a private clinical named entity recognition dataset owned by APHP. They also achieved similar scores to CamemBERT on EMEA and MEDLINE. Their new model performs better on clinical data, yet it obtains scores similar to CamemBERT in other biomedical domains.

\citetlanguageresource{labrak2023drbert} introduced a public French biomedical model named DrBERT. Through their experiments, they explored both continual-pretraining and from-scratch training strategies. Their findings indicated a superior performance when training from scratch, suggesting that continual-pretraining with CamemBERT for French biomedical data may not be as effective. However, when applied to PubMedBERT, continual-pretraining yielded results nearly on par with the from-scratch approach.

Finally, \citetlanguageresource{berhe:hal-03911564} introduced AliBERT, which was trained on a French biomedical corpus primarily comprising articles from ScienceDirect and theses collected through Sudoc. The model leverages a new regularized Unigram-based tokenizer and underwent extensive training on 48 GPUs for a total duration of 20 hours. Their pretrained model outperforms notable French non-domain-specific models, such as CamemBERT and FlauBERT, in two biomedical downstream tasks. Unfortunately, the model is not currently available.

\section{CamemBERT-bio}

\begin{table}[]\centering
\begin{tabularx}{\columnwidth}{ll >{\raggedleft\arraybackslash}Xr}\toprule
Corpus & Details &Size \\ \midrule
ISTEX &Scientific literature &276 M \\
CLEAR  &Drug leaflets &73 M \\
E3C  &  Clinical cases and leaflets  & 64 M \\
\bottomrule
Total & & 413 M \\
\end{tabularx}
\caption{Composition of the biomed-fr corpus (in millions of words)}\label{tab:biomed-fr}
\end{table}

\subsection{Corpus : biomed-fr}

First, we built a French biomedical corpus composed exclusively of public documents to minimize the usage constraints mentioned earlier. The documents come from three different sources (see Table.\ref{tab:biomed-fr}), the main one being ISTEX. This new corpus, named \textit{biomed-fr}, consists of 413 million words, equivalent to 2.7 GB of data. \citetlanguageresource{martin-etal-2020-camembert} have shown that with only 4 GB of data, it is possible to achieve performance almost comparable to the model trained with the 138 GB OSCAR dataset \citeplanguageresource{OrtizSuarezSagotRomary2019}. For an adaptation of CamemBERT to the biomedical domain, this amount of data can be considered sufficient.

\begin{table*}[]\centering
\small
\begin{tabular}{lllccc} \toprule
& & &  &\multicolumn{2}{c}{CamemBERT-bio} \\\cmidrule(lr){5-6}
Style &Dataset & Score & CamemBERT & \footnotesize{biomed-fr-small} & \footnotesize{biomed-fr} \\\midrule
\multirow{9}{*}{Clinical} &\multirow{3}{*}{CAS1} &F1 &70.50 $\pm$ 1.75 & \ul{72.94 $\pm$ 1.12} &\textbf{73.03 $\pm$ 1.29} \\
& &P &70.12 $\pm$ 1.93 &\textbf{72.97 $\pm$ 0.84} & \ul{71.71 $\pm$ 1.61} \\
& &R &70.89 $\pm$ 1.78 & \ul{72.92 $\pm$ 1.39} &\textbf{74.42 $\pm$ 1.49} \\
\cmidrule{2-6}
&\multirow{3}{*}{CAS2} &F1 &79.02 $\pm$ 0.92 & \ul{80.00 $\pm$ 0.32} &\textbf{81.66 $\pm$ 0.59} \\
& &P &77.3 $\pm$ 1.36 & \ul{78.29 $\pm$ 0.91} &\textbf{80.96 $\pm$ 0.91} \\
& &R &80.83 $\pm$ 0.96 & \ul{81.80 $\pm$ 0.48} &\textbf{82.37 $\pm$ 0.69} \\
\cmidrule{2-6}
\multirow{3}{*}{} &\multirow{3}{*}{E3C} &F1 &67.63 $\pm$ 1.45 & \ul{67.96 $\pm$ 1.85} &\textbf{69.85 $\pm$ 1.58} \\
& &P & \ul{78.19 $\pm$ 0.72} &77.41 $\pm$ 1.01 &\textbf{79.11 $\pm$ 0.42} \\
& &R &59.61 $\pm$ 2.25 & \ul{60.57 $\pm$ 2.32} &\textbf{62.56 $\pm$ 2.50} \\
\cmidrule{1-6}
\multirow{3}{*}{Leaflets} &\multirow{3}{*}{EMEA} &F1 &74.14 $\pm$ 1.95 & \ul{75.93 $\pm$ 2.42} &\textbf{76.71 $\pm$ 1.50} \\
& &P &74.62 $\pm$ 1.97 & \ul{76.23 $\pm$ 2.27} &\textbf{76.92 $\pm$ 1.96} \\
& &R &73.68 $\pm$ 2.22 & \ul{75.63 $\pm$ 2.61} &\textbf{76.52 $\pm$ 1.62} \\
\cmidrule{1-6}
\multirow{3}{*}{Scientific} &\multirow{3}{*}{MEDLINE} &F1 & \ul{65.73 $\pm$ 0.40} &65.48 $\pm$ 0.31 &\textbf{68.47 $\pm$ 0.54} \\
& &P & \ul{64.94 $\pm$ 0.82} &64.43 $\pm$ 0.50 &\textbf{67.77 $\pm$ 0.88} \\
& &R & \ul{66.56 $\pm$ 0.56} & \ul{66.56 $\pm$ 0.16} &\textbf{69.21 $\pm$ 1.32} \\
\bottomrule
\end{tabular}
\caption{F-scores on different biomedical named entity recognition tasks}\label{tab:bioVSbase}
\end{table*}

\paragraph{ISTEX} The ISTEX database contains references to 27 million scientific publications. We extracted 108,183 French documents published in a biology or medical journal since 1990. Articles published before this date often contain numerous typographical errors, as they are often scanned articles that require optical character recognition algorithms, resulting in a certain number of errors. Such errors are found to a lesser extent in articles published after 1990. Some documents, although in French, contain passages in English. Therefore, there is an indeterminate amount of English in this corpus. However, it is unlikely that this will significantly impact pre-training. Typographical errors and the presence of other languages are aspects that can be addressed in future versions of biomed-fr.

\paragraph{CLEAR} The CLEAR corpus \citeplanguageresource{grabar_clear_2018} consists of encyclopedia articles, drug leaflets, and abstracts of scientific articles. Each document is available in two versions: one in technical language and the other in simplified language. We retrieved all of these documents in both versions. Regarding the drug leaflets, we removed redundant sentences at the beginning and end of each document, such as the website navigation bar from which the documents were extracted or information about the company selling the documents.

\paragraph{E3C} This corpus \citeplanguageresource{magnini_e3c_2021} is composed of three layers. The first two layers are annotated or semi-annotated and will be used for evaluation. The last layer is not annotated, and that is the one we retrieved. It consists of medical specialty admission competitions, drug leaflets, and medical thesis abstracts. There may be duplicates of some leaflets found in the CLEAR corpus.

\paragraph{biomed-fr-small} By  randomly selecting 10\% of content from biomed-fr, we created a smaller corpus called \textit{biomed-fr-small}. The corpus allows us to study the impact of corpus size.

\subsection{Pre-training Strategies}

For the adaptation of CamemBERT to the biomedical domain, we conducted a second phase of pre-training on both versions of the biomed-fr corpus, starting from the weights and configuration of the camembert-base model. We applied the Masked Language Modeling (MLM) task with whole-word masking, following the method of \citetlanguageresource{martin-etal-2020-camembert}. We used the Adam optimizer \cite{kingma_adam_2017} with $\beta_1 = 0.9$ and $\beta_2 = 0.98$, and a learning rate of $5e-5$. We performed 50,000 steps over 39 hours using two Tesla V100 GPUs. A batch size of 8 per GPU and gradient accumulation over 16 steps were used to achieve an effective batch size of 256.

\subsection{Fine-tuning and Evaluation}

Regarding model evaluation, we collected three named entity recognition evaluation datasets. These datasets cover various styles, allowing us to assess the model's versatility across different subdomains of biomedicine.

\paragraph{QUAERO} The QUAERO corpus \citeplanguageresource{neveol14quaero} consists of two evaluation sets: EMEA, containing drug leaflets, and MEDLINE, containing scientific article titles. The entities are manually annotated following 10 semantic groups from the UMLS \citeplanguageresource{lindberg_unified_1993}. As some of these entities are nested, we kept only the entities with the coarsest granularity. F-scores are calculated in the same way.

\paragraph{E3C} For evaluation, unlike the biomed-fr corpus, we use layers 1 and 2. These layers contain documents of different types, including clinical cases extracted from scientific articles. Layer 2 is semi-annotated, and it is used as the training set for fine-tuning, with 10\% dedicated to the validation set. We evaluate on layer 1, which is fully manually annotated. There is only one class, and the objective is to find clinical entities in the text, regardless of their type.

\paragraph{CAS} The CAS corpus \citeplanguageresource{grouin_clinical_2019} also consists of clinical cases from scientific articles. We focus on task 3 of DEFT 2020 \cite{cardon_presentation_2020}, which is an information extraction task based on CAS. It includes two subtasks, and thus two sets of annotations. In the first subtask, two classes need to be identified: \textit{pathology} and \textit{signs or symptoms}. The second subtask concerns associated information, including \textit{anatomy}, \textit{dose}, \textit{examination}, \textit{mode}, \textit{timing}, \textit{substance}, \textit{treatment}, and \textit{value}. These two tasks will be referred to as CAS1 and CAS2, respectively.

\paragraph{Fine-tuning} For fine-tuning, we used Optuna \cite{akiba_optuna_2019} for hyperparameter selection. We set the learning rate to $5e-5$, the warmup ratio to 0.224, and the batch size to 16. We performed 2000 steps. Predictions were made using a simple linear layer on top of the model. None of the CamemBERT layers were frozen.

\paragraph{Evaluation} Scores are measured using the seqeval tool \cite{seqeval} in strict mode with micro-average and the "\textbf{IOB2}" scheme. For each evaluation, the best fine-tuned model on the validation set is selected to measure the final score on the test set. We average the results over 10 evaluations with different seeds.

\section{Results and Discussion}

\begin{table*}[h]\centering
\scriptsize
\begin{tabular}{llcccccccc}\toprule
& & \multicolumn{1}{c}{CAS1}  &\multicolumn{1}{c}{CAS2} &\multicolumn{3}{c}{EMEA} &\multicolumn{3}{c}{MEDLINE} \\ \cmidrule(lr){5-7}\cmidrule(lr){8-10}
Evaluator &Authors & \multicolumn{1}{c}{F1} & \multicolumn{1}{c}{F1} &\multicolumn{1}{c}{F1} &P &\multicolumn{1}{c}{R} &F1 &P &R \\\midrule
\multirow{4}{*}{seqeval} &\citet{dura_learning_2022}-fine-tuned &- &- & \ul{72.90} &- &- &59.70 &- &- \\ \cmidrule(lr){2-10}
&\citet{dura_learning_2022}-from-scratch &- &- &69.30 &- &- &\ul{60.10} &- &- \\ \cmidrule(lr){2-10}
&ours &\textbf{73.03} &\textbf{81.66} &\textbf{76.71} &\textbf{76.92} &\textbf{76.52} &\textbf{68.47} &\textbf{67.77} &\textbf{69.21}  \\ 
\multicolumn{10}{l}{}\\ 
\bottomrule
\multicolumn{10}{l}{}\\ 
\multirow{6}{*}{BRATeval} & \citet{le_clercq_de_lannoy_strategies_2022} & & &67.4 &\ul{73.4} &62.2 &55.3 &62.2 &\ul{49.7} \\ \cmidrule{2-10}
& \citet{mulligen_erasmus_2016} &- &- &\ul{74.9} &71.6 &\textbf{78.5} &\textbf{69.8} &\ul{68} &\textbf{71.6} \\ \cmidrule{2-10}
& \citet{copara_contextualized_2020} &\ul{61.53} &\ul{73.7} &- &- &- &- &- &- \\ \cmidrule{2-10}
&ours &\textbf{84.97} &\textbf{83.25} &\textbf{77.80} &\textbf{79.77} &\ul{75.93} &\ul{56.16} &\textbf{75.33} &44.82 \\
\bottomrule
\end{tabular}
\caption{Comparison of CamemBERT-bio with different approaches on the 4 named entity recognition tasks. In the first part of the table, scores are measured with seqeval \cite{seqeval}, and in the second part with BRATeval, which is the evaluation tool provided for the CLEF eHealth Evaluation lab 2016 campaign \cite{neveol_clinical_2016}.}
\label{tab:compare}
\end{table*}

\paragraph{CamemBERT vs CamemBERT-bio} We observe a significant performance gain on all evaluation datasets with our new model (see Table.\ref{tab:bioVSbase}). On average, we achieve a 2.54-point improvement in F-score. This gain is observed across all styles, demonstrating the model's versatility for both clinical and scientific domains.

\paragraph{biomed-fr-small vs biomed-fr} We observe a decrease in performance with the biomed-fr-small dataset, but there is still a significant gain on certain datasets compared to CamemBERT. This confirms that the size of the corpus positively influences the performance, even in a specialized domain like biomedicine.

\begin{table}[]\centering
\small
\begin{tabular}{lll}\toprule
&\multicolumn{2}{c}{Pre-training corpus} \\\cmidrule(){2-3}
Authors &Origin &Size\footnotemark[1] \\\midrule
\small{\citet{dura_learning_2022}} &APHP &21 MD \\
\scriptsize{\citet{le_clercq_de_lannoy_strategies_2022}} &misc &136 MW \\
\small{\citet{copara_contextualized_2020}} &PubMed &31 KD \\
ours & biomed-fr &413 MW \\ 
\bottomrule
\end{tabular}
\caption{Pre-training corpora of related works (cf. Table.\ref{tab:compare})} 
\label{tab:corpus}
\end{table}

% TODO: without afterpage the footnote is 2 page behind for some reason 
\afterpage{\afterpage{\footnotetext[1]{Units: MD (Million Documents), MW (Million Words), KD (Thousand Documents). As we rely on related articles, we can't provide a better estimation.}}}

\paragraph{Comparison with the state of the art} We compared the performance of CamemBERT-bio with various previously mentioned approaches (see Table.\ref{tab:compare}). CamemBERT-bio achieves the best results for almost all evaluation datasets. \citet{dura_learning_2022} did not observe improvement on EMEA and MEDLINE compared to CamemBERT because their pre-training corpus (see Table.\ref{tab:corpus}) consists of documents from APHP, making it a less diverse corpus. However, they gain several points on their evaluation dataset, which is also based on APHP documents. \citet{mulligen_erasmus_2016} presents the highest score on MEDLINE and the best recall on EMEA. Their approach is based on a knowledge-based model, allowing them to achieve the best recall on both QUAERO evaluation datasets. Furthermore, their approach is the only one in this table capable of handling nested entities, giving them an advantage.

It is important to note that these different CamemBERT-based approaches have various experimental setups. The presence of CRF layers instead of a simple linear layer after CamemBERT, freezing CamemBERT layers, and variations in hyperparameters are examples of differences observed in addition to the pre-training corpus, which makes the comparison more challenging.

\paragraph{Tokenization analysis} CamemBERT-bio is a biomedical-adapted model based on CamemBERT. Unlike a new model trained from scratch, it shares the same vocabulary. The vocabulary of CamemBERT was constructed using SentencePiece \citeplanguageresource{kudo_sentencepiece_2018} on an OSCAR sample. Therefore, it is a general-purpose vocabulary designed for everyday language. We can hypothesize that the tokenization of CamemBERT may result in oversplitting of biomedical technical terms.

To investigate this possibility, we trained a specialized tokenizer on biomed-fr-small and calculated the intersection of the two vocabularies (Table.\ref{tab:tok}).

\begin{table*}[]\centering

\begin{tabular}{l@{\hskip 2cm}c@{\hskip 1cm}cc}\toprule
Terms &general &specialized \\ \midrule
échocardiographie &écho-cardi-ographie &échocardiographi-e \\
transthoracique &trans-thorac-ique &trans-thoracique \\
glimépiride &g-lim-épi-ride & gli-m-épi-ride \\
cardiopathie & cardio-pathie & cardiopathie \\
diastoliques & dia-s-tol-iques & diastolique-s \\
\bottomrule
\end{tabular}
\caption{Comparison of tokenization between a general-purpose tokenizer and a specialized tokenizer for some biomedical technical terms.}\label{tab:tok}
\end{table*}

We find a 45\% intersection between the two vocabularies, which is quite close to the 42\% intersection found by \citetlanguageresource{beltagy_scibert_2019} between the vocabulary of BERT and SciBERT. Therefore, there is a significant difference in the most frequent terms.

\section{Evaluating Models: Methodology and Discussion}

In a recent work, \citetlanguageresource{labrak2023drbert} introduced a new French biomedical language model named DrBERT. The authors contend that performing continual-pretraining on biomedical data from CamemBERT leads to reduced performance. They used a different methodology for the evaluation of named entity recognition, prompting us to examine the implications of each approach on the interpretation of the results.

Our evaluation approach is centered around micro F1, which is measured using seqeval in strict mode with the IOB2 scheme. Their approch is based on weighted F1 with independant token classification. Every token has a label and the model is evaluated for each of them, whereas with seqeval, the model is evaluated for each entity. Notably, the "O" token, representing non-entity, is the most frequent token and thus holds significant weight in the evaluation process. As all tokens are labeled, the number of predictions depends on the tokenizer, thereby influencing the final score.  In order to explore the impact of token labeling variations on the evaluation,  we conducted an additional experiment on EMEA and MEDLINE where we reproduced the DrBERT methodolody that we named \textit{token-with-O}, and another one where we excluded the "O" token and only labeled the first token of each entity, that will be refered as \textit{entity-without-O}. Furthermore, we included a seqeval strict score using the IOB2 scheme, although it is not directly comparable to our approach due to their implementation of nested entities by concatenating their names to form new entities, while we simply removed them.

\begin{table*}[]\centering
\scriptsize
\begin{tabular}{llcccccccc}\toprule
& &\multicolumn{4}{c}{EMEA} &\multicolumn{4}{c}{MEDLINE} \\ \cmidrule(lr){3-6}\cmidrule(lr){7-10}
Methodology &Model &\multicolumn{1}{c}{weighted-f1} &macro-f1 &\multicolumn{1}{c}{micro-f1} & seqeval-f1 &weighted-f1 &macro-f1 &micro-f1 & seqeval-f1 \\\midrule
\multirow{5}{*}{\textit{token-with-O}}
& DrBERT-7GB & 87.45 & 34.95 & - & - & 75.52 & \textbf{15.07} & - & - \\ \cmidrule(lr){2-10}
& CamemBERT-bio & \textbf{90.37} & \ul{36.27} & - & - & \textbf{77.89} & \ul{14.82} & - & - \\ \cmidrule(lr){2-10}
& CamemBERT & \ul{88.33} & \textbf{47.45} & - & - & \ul{76.2} & 11.92 & - & - \\ 
\multicolumn{10}{l}{}\\ 
\bottomrule
\multicolumn{10}{l}{}\\ 
\multirow{4}{*}{\textit{entity-without-O}} 
& DrBERT-7GB & 66.72 & \textbf{24.72} & 68.34 & 59.39 & 60.70 & \textbf{10.80} & 63.40 & 50.45 \\ \cmidrule(lr){2-10}
& CamemBERT-bio & \textbf{73.53} & \ul{24.15} & \textbf{75.05} & \textbf{67.58} & \textbf{62.04} & 8.695 & \textbf{65.44} & \textbf{52.9} \\ \cmidrule(lr){2-10}
& CamemBERT & \ul{71.85} & 22.71 & \ul{72.93} & \ul{64.23} & \ul{60.95} & \ul{9.413} & \ul{63.47} & \ul{51.75} \\ 
\bottomrule
\end{tabular}
\caption{Performance comparison of CamemBERT, CamemBERT-bio, and DrBERT on EMEA and MEDLINE using the evaluation methodology \textit{token-with-O}, along with a modified variant \textit{entity-without-O}. The reported scores are averaged over 10 runs.}
\label{tab:compare-meth}
\end{table*}

We observe significant differences in scores between the two methodologies (Table.\ref{tab:compare-meth}). The methodology \textit{token-with-O} consistently reports higher scores across all tasks compared to the alternative method, \textit{entity-without-O}. However, the reported scores do not precisely align with those presented by \citetlanguageresource {labrak2023drbert}, where DrBERT exhibits improvement over CamemBERT for EMEA, and match CamemBERT for MEDLINE. We posit that these discrepancies arise from disparities in hyperparameters. Nevertheless, the comparison with \textit{entity-without-O} still provides insightful observations.

Notably, the best-performing model varies depending on the chosen methodology for one metric. CamemBERT achieves the highest macro F1 score on the EMEA dataset, whereas with \textit{entity-without-O}, DrBERT-7GB emerges as the top model. This observation underscores the influential role of the "O" class in determining the best-performing model and prompts us to consider the aspect we aim to evaluate.

In terms of macro F1, DrBERT consistently outperforms CamemBERT-bio when evaluated using the \textit{entity-without-O} methodology. On the other hand, CamemBERT-bio exhibits better performance across all other metrics. This indicates that DrBERT demonstrates a more balanced performance across different classes. These findings suggest that these models may possess complementary strengths, and it is important to focus on enhancing the performance of CamemBERT-bio in this aspect.

It is worth noting that the evaluations conducted by \citetlanguageresource{labrak2023drbert} encompass a wide range of tasks beyond EMEA and MEDLINE. This observation underscores the importance of establishing a unified benchmark to facilitate fair comparisons between models for the french biomedical domain. A noteworthy tool for this purpose is BRATeval from the CLEF eHealth Evaluation lab 2016 campaign \cite{neveol_clinical_2016}, as it evaluates based on exact match character offsets, thus avoiding any influence from dataset pre-processing methods.

Furthermore, \citetlanguageresource{labrak2023drbert} suggested that continual-pretraining from a French model on French biomedical data isn't effective, as it results in an F1-score loss of up to 20 points. However, considering our success with continual-pretraining and the points we raised about the methodology, we suggest a reassessment of these findings.

\section{Environmental Impact}

Considering the environmental implications is crucial when discussing language model training due to its potential compute intensity.

\begin{table*}[]\centering
\small
\begin{tabular}{lllll}
\toprule
 & Training time (hours) & Hardware type & Total GPU-hours & \begin{tabular}[c]{@{}l@{}}Estimation of \\ carbon emitted (kg CO2 eq.)\end{tabular} \\ \midrule
DrBERT & 20h & 128xV100 & 2560 & 26.11 \\  \midrule
AliBERT & 20h & 48xA100 & 960 & 8.16 \\  \midrule
CamemBERT-bio & 39h & 2xV100 & 78 & 0.8 \\ \bottomrule
\end{tabular}
\label{tab:carbon}
\caption{Carbon emitted estimation based on hardware and training time. We used a rate of 34g CO2eq. per kWh, reflecting the average over the last 12 months in France starting from September 2022. This time frame and location coincide with when and where all experiments were conducted.}
\end{table*}

Considering estimated carbon emissions \footnote{Estimations were conducted using the \href{https://mlco2.github.io/impact}{MachineLearning Impact calculator} presented by \citet{lacoste2019quantifying}} during training, AliBERT emits 10 times the amount of CamemBERT-bio, while DrBERT releases 32 times more (Table.\ref{tab:carbon}). While a direct comparison with AliBERT wasn't possible, the minor performance variance with DrBERT, set against the pronounced disparities in computational and environmental costs, leads us to advocate for continual-pretraining as the preferred adaptation method for biomedical language models.

\section{Limitations}

It is important to discuss some limitations of our studies. 

Firstly, our training corpus \textit{biomed-fr} is limited in its diversity. This is because we chose to only use publicly available materials, which tend to lean towards scientific content. As a result, our analysis may not fully represent performances on real private clinical data.

Furthermore, we didn't explore any potential biases in our data and some part of our dataset would benefit from further cleaning. These aspects could impact the outputs of our model and should be investigated.

Additionally, our evaluation focused on one task which is Named Entity Recognition. This might mean our understanding of how well our model performs is restricted. Future studies should aim to assess our model across a wider range of tasks and datasets to get a clearer picture of its strengths and weaknesses.

This underscores the need for further research to address these constraints and enhance our understanding of the subject matter.

\section{Conclusion and Perspectives}

We have introduced a new French biomedical corpus called \textit{biomed-fr} consisting of 413 million words, composed of drug leaflets and documents from scientific literature in medicine and biology. This new corpus has allowed us to adapt CamemBERT to the biomedical domain through a second phase of pre-training. We observe an improvement in performance on all our named entity recognition evaluation datasets, with an average gain of 2.54 F-score points. Our model establishes a new state-of-the-art on these French biomedical language processing tasks.
We have some directions for future versions of \textit{biomed-fr}. Firstly, we can further clean the data by removing passages within the documents that are not in French or by excluding documents with a high number of typographical errors. Secondly, we can increase the amount of data. This could involve leveraging archived documents on HAL related to life sciences, particularly those published by INSERM, or retrieving abstracts of French articles from PubMed.

The analysis of tokenization prompts us to consider expanding the vocabulary for CamemBERT-bio. The relatively modest performance gain of PubMedBERT compared to BioBERT and the similar performance of DrBERT compared to CamemBERT-bio despite their specialized vocabulary, the over-segmentation of technical terms and the low intersection rate between the generalist vocabulary and the specialized vocabulary demonstrate the value of the experiment. However, taking into account the comparable performance levels and significant difference in environmental impact, we advocate strongly for the continual-pretraining method in adapting language models to the biomedical domain.

Finally, in recent months, numerous generative models, often with billions of parameters, have demonstrated remarkable performance on biomedical tasks, sometimes surpassing specialized models like BioBERT \cite{agrawal_large_2022, singhal_large_2022}. This is a promising research direction for biomedical information extraction. However, we have reasons to believe that BERT-type models still have value \cite{lehman_we_2023}. Firstly, in a clinical context, models are often used within healthcare institution infrastructures, which entails resource constraints. It is easier to deploy small specialized models than large generalist models in such cases. Secondly, the use of these generative models often requires accessing remote servers, typically through APIs, which makes their utilization challenging considering the confidentiality constraints imposed on clinical documents.

%\section*{Acknowledgements}
%We want to warmly thank Wissam Antoun, Roman %Castagné, Benoît Sagot and Djamé Seddah for %their comments and their methodological %insights. The authors are grateful to the %CLEPS infrastructure from the Inria of Paris %for providing resources and support. 


% uncommenting the following lines
% introduces many non-relevant entries
%\nocite{*}
\section{Bibliographical References}\label{sec:reference}

\bibliographystyle{lrec-coling2024-natbib}
\bibliography{ref}

\section{Language Resource References}
\label{lr:ref}
\bibliographystylelanguageresource{lrec-coling2024-natbib}
\bibliographylanguageresource{languageresource}

\begin{table*}[htp!]
\small
\centering
\begin{tabular}{lllccc} \toprule
& & &  &\multicolumn{2}{c}{CamemBERT-bio} \\\cmidrule(lr){5-6}
Style &Dataset & Score & CamemBERT & \footnotesize{biomed-fr-small} & \footnotesize{biomed-fr} \\\midrule
\multirow{6}{*}{Clinical} &\multirow{3}{*}{CAS1} &F1 &67.85 & \ul{70.18} &\textbf{71.37} \\
& &P &73.13 & \ul{74.11} & \textbf{75.22} \\
& &R &64.38 & \ul{67.39} &\textbf{68.34} \\
\cmidrule{2-6}
&\multirow{3}{*}{CAS2} &F1 &72.40 & \textbf{74.69} &\ul{74.32} \\
& &P &73.22 & \textbf{74.46} &\ul{72.46} \\
& &R &72.03& \ul{75.23} &\textbf{77.56} \\
\cmidrule{2-6}
\multirow{3}{*}{Leaflets} &\multirow{3}{*}{EMEA} &F1 &50.74 & \ul{53.1} &\textbf{55.69} \\
& &P &51.99 & \ul{55.88} &\textbf{56.01} \\
& &R &50.67 & \ul{51.78} &\textbf{56.09} \\
\cmidrule{1-6}
\multirow{3}{*}{Scientific} &\multirow{3}{*}{MEDLINE} &F1 & 45.09 &\ul{47.73} &\textbf{48.18} \\
& &P & 46.57 & \ul{48.04} &\textbf{49.16} \\
& &R & 47.18 & \textbf{52.62} &\ul{50.38} \\
\bottomrule
\end{tabular}
\caption{F-scores on different biomedical named entity recognition tasks with macro-average}\label{tab:macro-bioVSbase}
\end{table*}
\section*{Appendices}

\newpage

\addcontentsline{toc}{section}{Appendices}
\renewcommand{\thesubsection}{\Alph{subsection}}

\subsection{Macro-average}

In our main results (Table.\ref{tab:bioVSbase}), scores are measured using the seqeval tool \cite{seqeval} in strict mode with micro-average and the "\textbf{IOB2}" scheme. We also conducted the same evaluation using macro-average (Table.\ref{tab:macro-bioVSbase}).

\end{document}